\documentclass[letterpaper, 10 pt, conference]{ieeeconf}  %

\IEEEoverridecommandlockouts                              %
\usepackage[T1]{fontenc}                                                          

\usepackage{cite}
\usepackage{hyperref}
\hypersetup{
    colorlinks=true,
    linkcolor=blue,
    filecolor=magenta,      
    urlcolor=cyan,
    pdftitle={Social 3D Scene Graphs: Modeling Human Actions and Relations for Interactive Service Robots},
    pdfpagemode=FullScreen,
    }
\usepackage{amsmath,amssymb,amsfonts}
\usepackage{algorithmic}
\usepackage{graphicx}
\usepackage{textcomp}
\usepackage{xcolor}
\usepackage{booktabs}
\usepackage{comment}
\usepackage{caption}
\usepackage{tikz}
\usetikzlibrary{matrix,calc}
\usepackage[table]{xcolor}
\usepackage{multirow}
\definecolor{neongreen}{RGB}{57, 255, 20}
\definecolor{neonorange}{RGB}{255, 95, 31}

\newcommand{\datasetname}{SocialGraph3D}
\newcommand{\modelname}{ReaSoN}
\newcommand{\modelfullname}{RElationship and SOcial reasoNing in 3D}
\newcommand{\cgname}{CGH}
\newcommand{\cgfullname}{ConceptGraphH}
\newcommand{\ablation}{\modelname{}$_{\text{w/o BD}}$}
\newcommand{\m}[1]{\mathcal{#1}}
\newcommand{\reasoningtask}{Social Scene Understanding~}

\title{\LARGE \bf
Social 3D Scene Graphs: Modeling Human\\Actions and Relations for Interactive Service Robots

}
\author{%
  Ermanno Bartoli$^{1*}$ \quad
  Dennis Rotondi$^{2,3*}$ \quad
  Buwei He$^{1}$ \quad
  Patric Jensfelt$^{1}$ \quad
  Kai O. Arras$^{2\dagger}$ \quad
  Iolanda Leite$^{1\dagger}$ \\
  \thanks{$^{*}$Equal contribution. $^{\dagger}$Equal supervision.}%
  \thanks{This work was partially funded by grants from the Swedish Research Council (2024-05867), the Swedish Foundation for Strategic Research (SSF FFL18-0199), the Digital Futures research center, the Vinnova Competence Center for Trustworthy Edge Computing Systems and Applications at KTH, and the Wallenberg Al, Autonomous Systems and Software Program (WASP) funded by the Knut and Alice Wallenberg Foundation.}%
  \thanks{$^{1}$E. Bartoli, B. He, P. Jensfelt and I. Leite are with Faculty of Robotics Perception and Learning, KTH Royal Institute of Technology, Stockholm, Sweden. Email: bartoli@kth.se}%
  \thanks{$^{2}$D. Rotondi and K.O. Arras are with the Socially Intelligent Robotics Lab at the Institute for Artificial Intelligence, University of Stuttgart, Germany. Email: dennis.rotondi@ki.uni-stuttgart.de}%
  \thanks{$^{3}$D. Rotondi is also part of the International Max Planck Research School for Intelligent Systems (IMPRS-IS).}
}

\newcommand{\acceptance}{%
\begin{tikzpicture}[overlay, remember picture]
\path (current page.north) ++(0,-1cm) node[align=center] {
This paper has been accepted for publication at the 2026 International Conference on Intelligent Robots and Systems (IROS).
};
\end{tikzpicture}
}
\newcommand{\isArxiv}[2]{#1} %

\begin{document}
\makeatletter

\maketitle
\isArxiv{\acceptance}{}

\begin{abstract}

Understanding how people interact with their surroundings and each other is essential for enabling robots to act in socially compliant and context-aware ways. %
While 3D Scene Graphs have emerged as a powerful semantic representation for scene understanding, existing approaches largely ignore humans in the scene, also due to the lack of annotated human-environment relationships. 
Moreover, existing methods typically capture only open-vocabulary relations from single image frames, which limits their ability to model long-range interactions beyond the observed content. 
We introduce Social 3D Scene Graphs, an augmented 3D Scene Graph representation that captures humans, their attributes, activities and relationships in the environment, both local and remote, using an open-vocabulary framework.
Furthermore, we introduce a new benchmark consisting of synthetic 
and real
environments with comprehensive human-scene relationship annotations and diverse types of queries for evaluating social scene understanding in 3D. 
The experiments demonstrate that our representation improves human activity prediction and reasoning about human-environment relations, paving the way toward socially intelligent robots.
\end{abstract}

\section{Introduction}
Service robots are increasingly being deployed in human-centric environments such as homes and offices.
In order to be effective in such spaces, they need more than just metric representations \cite{Cadena_2016}. 
They must also understand how humans interact with their surroundings and reason about what these interactions entail, e.g., to provide the right service at the right time. 
For example, a person sitting on a sofa and watching television defines a leisure zone where the robot should avoid blocking the view. 
A person at a dining table with an empty glass creates a meal context where the robot should infer the need for water rather than beginning to clean.
Social robots must be able to accurately interpret such situational cues to achieve socially compliant, anticipatory, and context-aware behavior.
\begin{figure}[t!]
\centering
\begin{minipage}{0.485\textwidth}
  \centering
  \includegraphics[width=1\linewidth,
  height=0.4\textheight
  , trim=0cm 6cm 0cm 2mm, clip]{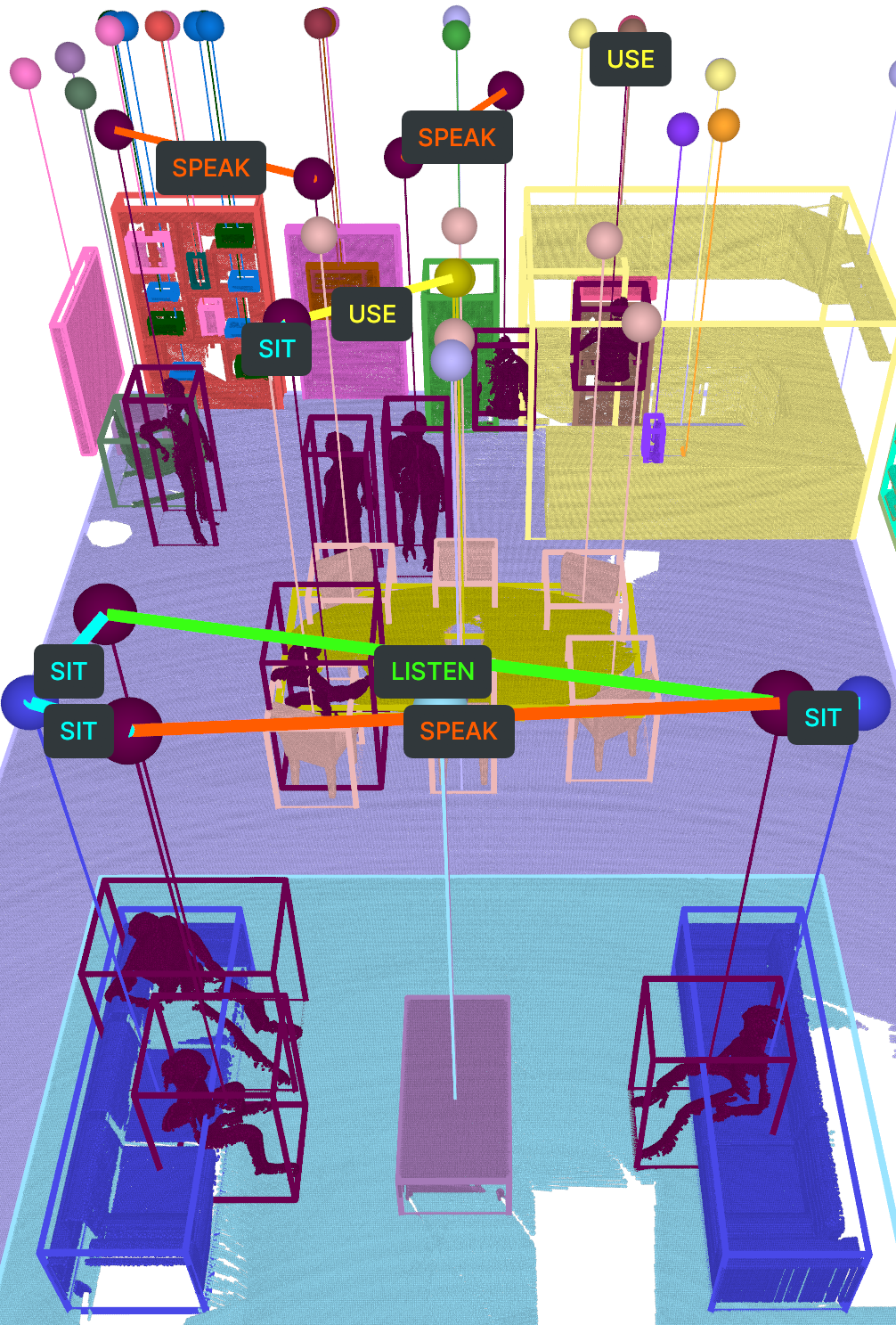}
    \caption{Example of a Social 3D Scene Graph. Each entity is represented as a node in the graph and visualized as a sphere. These nodes are connected by social edges (e.g., \textcolor{neonorange}{\texttt{SPEAK}} and \textcolor{neongreen}{\texttt{LISTEN}}) that capture their activities in the scene.
    }
    \label{fig:social3dsg}
    \vspace{-16pt}
\end{minipage}%
\end{figure}

3D Scene Graphs (3DSGs) \cite{rotondi20263dscenegraphsopen, armeni20193dscenegraphstructure, Kim_2020} offer a promising environment representation for robots to achieve these goals.
They have recently gained popularity in the robotics community as a semantic representation for scene understanding and downstream tasks, due to their ability to compactly encode rich 3D information \cite{hughes2024foundations} and their versatility in a wide range of applications.
However, a major limitation of current 3DSG approaches is the absence of human relationships within the 3D environments where they are deployed~(e.g., \cite{dai2017scannet, replica19arxiv, baruch2021arkitscenes, yeshwanth2023scannet, zhang2025open}). 
This is largely due to the scarcity of annotated data capturing human-environment relationships, which led researchers to use non-public datasets for evaluating their methods involving humans~\cite{rosinol2021kimera}. 
In addition, state-of-the-art 3DSG methods typically capture open-vocabulary spatial relationships only at a local level, extracting them from a single view by prompting a vision-language model (VLM) or by comparing 3D features such as embeddings or occupied volume. 
Yet, humans frequently interact with objects or other people across larger spatial extents, beyond the scope of a single frame%
, making standard 3DSG construction algorithms unsuited for modeling such interactions.

Motivated by these challenges, we address the problem of generating Social 3D Scene Graphs (S3DSG), i.e., 3DSGs that model human activities and their relationships with the environment beyond local single-frame interactions, by assuming that humans look at the objects they engage with (see Fig.~\ref{fig:social3dsg}).
Specifically, our contributions are:
\begin{itemize}
    \item We introduce Social 3D Scene Graphs and propose a method to build them, augmenting 3DSGs with humans' relationships and activities through our  \modelname{} module, which operates in open-vocabulary settings. %

    \item We introduce a benchmark, \datasetname, to evaluate human relationships, activities and social scene understanding in 3D. 
    We open-source our \href{https://github.com/Ermanno-Bartoli/Social-3D-Scene-Graphs}{code and dataset}.

    \item We use our benchmark to demonstrate that our representation outperform state-of-the-art performance on the introduced tasks, and we qualitatively show how S3DSG can be applied to a real-world downstream task.
\end{itemize}

\section{Related Work}
\subsection{3D Scene Graphs}\label{sec:related_work_3dsg}
Armeni et al.~\cite{armeni20193dscenegraphstructure} introduced the concept of 3D Scene Graphs, \(\m{G} = (\m{V}, \m{E})\), a hierarchical structure built over a 3D representation such as a mesh or point cloud, in which nodes \(\m{V}\) represent objects and spatial entities (e.g., rooms, floors, buildings), and edges \(\m{E}\) capture relationships between them, including spatial (e.g., A ``on top of'' B), comparative (e.g., A ``smaller than'' B), and support (e.g., A ``standing on'' B).

Follow-up works~\cite{hydra, clio, conceptgraph, hov-sg, rotondi2025fungraph, werby2025keys} aimed to construct such representations from sets of registered RGB-D images. 
However, these methods focus exclusively on spatial relationships inferable from a single image via a VLM. By requiring objects to appear within the same frame, they overlook human behavioral patterns: specifically, the tendency to interact with distant objects across a broader representation.

In parallel, the vision community has explored the task of predicting 3D semantic scene graphs~\cite{wald2020learning3dsemanticscene, wu2021scenegraphfusionincremental3dscene, zhang2021knowledge, wang2023vlsatvisuallinguisticsemanticsassisted, koch2024open3dsgopenvocabulary3dscene} from class-agnostic segmented point clouds, framing it as a classification problem typically addressed using iterative message-passing schemes~\cite{xu2017scenegraphgenerationiterative} with graph neural networks, which however suffer from the vanishing gradient problem \cite{pascanu2013difficulty} and therefore often restrict the number of iterations, limiting their ability to capture long-range relationships.
While the predicted relationships are generally not constrained to the locality of a single image frame, these approaches focus solely on 3D features and are trained on the 3DSSG dataset~\cite{wald2020learning3dsemanticscene}, which, like the dataset from~\cite{armeni20193dscenegraphstructure}, is limited to close-vocabulary object-centric relationships.

Zhang et al.~\cite{zhang2025open} introduce the dichotomy of local and remote relationships and take an initial step toward addressing open-vocabulary remote functional relationships by proposing a confidence-aware reasoning approach. This method leverages the commonsense knowledge of a large language model to infer links between functionally related but spatially distant objects, such as a light switch and the light it controls. 
However, their framework remains object-centric, focusing solely on functional relations and neglecting humans entirely.

\subsection{Humans in 3D Scene Graphs}
Motivated by the need to track agents in a scene, 3D Dynamic Scene Graphs~\cite{rosinol20203ddynamicscenegraphs, rosinol2021kimera, ravichandran2022hierarchical} model the temporal evolution of human poses as nodes and generate mesh representations of them; however, they capture only relationships between places and pairwise spatio-temporal relations (e.g., “agent A is in room B at time t”). 
The data in these works come from a Unity simulator, but most resources, such as the semantic mesh, are not publicly available.
In contrast, our work builds and publicly releases scenes with synthetic and real humans, complete with full annotations.

Building on 3D dynamic scene graphs, the work most closely related to ours is that of Gorlo et al.~\cite{gorlo2024longtermhumantrajectoryprediction}, who tackle the problem of estimating human trajectories in complex environments by reasoning about human-scene interactions. In their approach, however, interactions with the environment are used solely to model how individuals navigate their surroundings. Furthermore, their method makes the strong assumption that only a single person is present in the scene, thereby excluding by design all interactions between people 
and reducing the dynamic tracking problem to a detection problem. 
In contrast, we explicitly model such interactions, which we consider crucial for a robot's contextual understanding: for example, recognizing that it should not disturb people who are engaged in conversation.

\subsection{Planning with 3D Scene Graphs}
\begin{figure*}[ht!]
\centering
  \centering
  \includegraphics[width=0.9\linewidth]{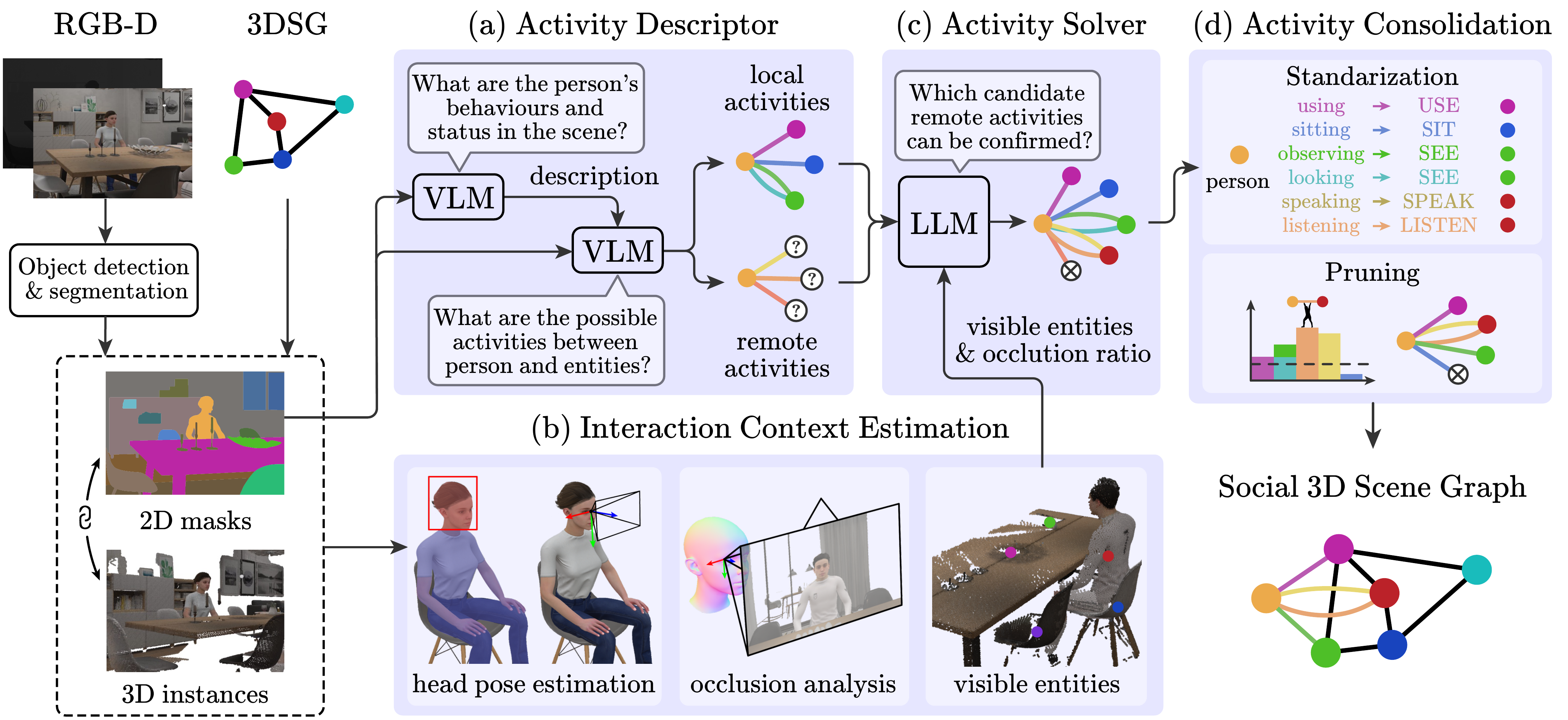}
    \caption{An overview of our solution \modelname{}, which extends existing 3D Scene Graphs with humans and their activities through four key components: \textbf{(a) Activity Descriptor} segments humans and generates behavior descriptions using VLMs, and identifies both local and remote activities under the context; \textbf{(b) Interaction Context Estimator} estimates head pose and models each person's visible frustum to determine visible entities through occlusion analysis; \textbf{(c) Activity Solver} uses LLM spatial reasoning to validate remote activities by finding relevant entities; and \textbf{(d) Activity Consolidation} refines activities using semantic frames and prunes spurious detections based on frequency thresholds, producing a robust Social 3DSG.}
    \label{fig:method}
    \vspace{-12pt}
\end{figure*}
3DSGs are well-suited for robot navigation, task planning, and motion planning because they provide topological environment models that are easily compatible with classical PDDL solvers~\cite{agia2022taskography, talukder2024anticipatoryplanning} and can support reasoning with LLMs~\cite{rana2023sayplan, liu2025delta, bartoli2025}.
For example, consider the task: “A postdoc spilled their soda. Help them clean it up.” used in \cite{rana2023sayplan}, where such queries are employed for planning. 
In contrast, our goal is to embed this knowledge directly into the representation: by explicitly modeling entities such as the postdoc and the soda, the representation enables an LLM to reason about the situation, for example, by deciding whether to avoid the spill or clean it up.

Additionally, 3DSGs have proven effective for grounding navigation goals expressed in unconstrained natural language~\cite{rajvanshi2024saynav, yin2024sg}. By modeling social relationships, it becomes possible to incorporate costs and clearance into the navigation process, enabling socially aware behaviors such as avoiding passing in front of people watching TV.

We share similar goals with the field of social robot navigation, which focuses on learning motion behaviors for human environments that are safe, efficient, and considered socially acceptable \cite{mavrogiannis2023core}. Examples include group-aware avoidance behaviors that use a \emph{social graph} whose edge costs are learned to encode group affiliation \cite{7487452}, approach behaviors to groups of humans for multi-party interaction \cite{sisbot2007human}, or predictive motion planning in dense crowds \cite{heuer2023proactive}. 
While these works can handle dynamic scenes, they are typically constrained to 2D motion behavior and offer limited social and semantic scene understanding. 
Our 3DSG approach aims to support such tasks with rich 3D semantic representations and enable human-centric reasoning and queries.

\section{Method}
\subsection{Problem Formulation}
Our objective is to build Social 3D Scene Graphs, enabling robots to understand and operate in human-centric environments more effectively.
We define a Social 3D Scene Graph as an extension of a traditional 3D Scene Graph that incorporates nodes and edges representing humans and their activities in addition to objects and spatial relationships.
Formally, a Social 3D Scene Graph is represented as a tuple $\m{G} = (\m{V}, \m{E})$ where:
\begin{itemize}
    \item $\m{V} = \m{V}_o \cup \m{V}_h$ is the set of nodes, where $\m{V}_o$ represents objects and $\m{V}_h$ represents humans in the environment.
    \item $\m{E} = \m{E}_s \cup \m{E}_a$ is the set of edges, where $\m{E}_s$ represents spatial relationships and $\m{E}_a$ represents activity relationships (e.g., ``sitting", ``talking to", ``looking at").
\end{itemize}

Our goal is not to reconstruct the complete 3D representation of the environment, but rather to incorporate humans and their activities into an existing 3DSG $\mathcal{SG}$.
We assume the availability of $\mathcal{SG}$ for an environment (without human representations), grounded to a 3D point cloud and derivable using methods such as ConceptGraphs~\cite{conceptgraph}.
The challenge lies in accurately detecting humans, estimating their interaction field, and inferring their activities from a sequence of registered RGB-D observations $\mathcal{I} = \{I_1, I_2, ..., I_n\}$ of the same scene and coordinate system of $\mathcal{SG}$. 
We assume that in these images, the humans remain still except for facial expressions and mouth movements.
We believe this approach is particularly relevant for service robots operating in human environments, where the primary sources of change are the humans and their activities, the surroundings remain static and objects are assumed not to move.
However, by handling the reconstruction and processing of humans at the final stage of a 3D Scene Graph generation pipeline, our approach can be seamlessly integrated into existing systems. 
We refer to our solution as the \modelname{}{} module.

\subsection{\modelfullname{}}
In this section, we describe how our module ``\modelfullname{}" (\modelname{}) augments a 3D Scene Graph with human-related information to construct Social 3D Scene Graphs.
As illustrated in Fig.~\ref{fig:method}, \modelname{} is composed of four submodules and activates whenever at least one human is detected. 
The pipeline processes each RGB-D observation $I_k \in \mathcal{I}$ by (i) detecting objects and humans (collectively referred to as entities), (ii) aligning detected objects with existing nodes in $\mathcal{SG}$, and (iii) estimating their activities through an understanding of fields of view, body pose, and socially relevant attributes.
The resulting graph is an augmented version of $\mathcal{SG}$ that explicitly encodes human behaviors and interactions.

\textbf{Activity Descriptor}.  
This module takes the entity detections as input and generates candidate descriptions of human behaviors along with their potential relationships.
Human bounding boxes are used as prompts for a segmentation model to generate masks.
We then apply morphological dilation to the resulting masks to highlight the contours and visually label the people in the image with their respective IDs and markers.
As shown in~\cite{fangandliu2024moka}, the markers enhance the generalization ability of the VLM, which analyzes the annotated image and outputs, for each human, a detailed characterization of their behaviors, including \textit{posture} (e.g., upright, leaning), \textit{gaze} (e.g., direct, averted), \textit{physical state} (e.g., standing, sitting), and other socially relevant attributes.
We then re-annotate the original input image $I_k$, labeling not only the humans but also all detected objects in the scene.
The VLM takes as input the newly annotated image and the behavior descriptions from the previous step, and infers the activities the humans are or might be performing.
This process yields two sets of relationships:
\begin{itemize}
\item \textit{Local activities}: activities for which the relevant object or person involved is present in the image, allowing the VLM to confidently confirm their occurrence.
\item \textit{Remote activities}: activities that might be inferred from the behavior description and the image, but cannot be confidently confirmed due to insufficient context (e.g., the relevant object or person is outside the image).
\end{itemize}

\textbf{Interaction Context Estimator.}
The interaction context estimator is introduced to determine the entities visible from a person's perspective, thereby enabling the understanding of remote activities beyond the directly observable field of view of a single image. 
The estimator focuses on two key aspects: (i) head pose estimation, used to approximate each person’s visual perceptual field, and (ii) occlusion analysis, which identifies the visually perceivable entities for each person along with their occlusion ratios.

For head pose estimation, we employ a two-stage strategy: person heads are first localized using a pre-trained detector, after which their full-range 3D orientations are estimated with a pre-trained model, and the head centroid is obtained from the depth image to anchor orientation in 3D space.

Given the estimated head pose, we model each person's visual frustum and render scene entities from the corresponding viewpoint as depth images. To address sparsity in the projected point clouds, we interpolate depth values within each entity's silhouette to obtain a dense depth proxy. Occlusions are then resolved through per-pixel depth comparisons, and the visibility of each entity is quantified as the fraction of its projected silhouette that remains visible, following the silhouette occlusion index (SOI)~\cite{soi}.

\textbf{Activity Solver.} 
The central objective here is to identify the entities with which to establish relationships within the set of visible ones, validating remote activities through the emerging spatial reasoning capabilities of LLMs~\cite{cheng2025spatialrgpt}.
For a set of remote activities identified as requiring disambiguation and their human behavior descriptions provided by the Activity Descriptor, the Activity Solver inspects the partial 3D scene graph that includes the entities within the human's field of view to locate relevant entities. 
For example, if a person appears to be reading but no book is directly visible, the VLM searches for plausible reading materials within the interaction space, taking into account the visible entities and their occlusion ratio from each person's view, and typical reading distances.

\textbf{Activity consolidation.}
After all images are processed, activity consolidation finalizes the Social 3D Scene Graph by refining activities for coherence and robustness.
For each human-entity pair connected by at least one edge, we ensure that all activity labels are clustered within augmented semantic frames that capture the full range of linguistic synset variation~\cite{di2019verbatlas, navigli2024nounatlas}.
For example, for the pair (Person1, Person2), activities such as “chatting with,” “speaking to,” and “talking to” are all canonicalized under the semantic frame \texttt{SPEAK}.
This consolidation ensures that a single, consistent representation is used for each unique activity type throughout the 3D Scene Graph, while preserving the open-vocabulary capabilities of the method.
We keep track of the number of activities clustered in each frame, which informs the subsequent pruning step.
To minimize errors in the pipeline, such as visual ambiguities, particularly within the Activity Descriptor module, we prune activities based on their detection frequency.
The key idea is that correct activities will appear more consistently across observations than spurious ones.

An edge $e \in \m{E}$ is retained only if its detection count $N(e)$ satisfies both relative and absolute thresholds:

\begin{equation}
\label{eq:pruning_condition}
\text{Keep}(e) \Longleftrightarrow N(e) \geq \max(\tau \cdot M, N_{\text{min}})
\end{equation}

where $M = \max_{e'} N(e')$ is the maximum count among edges from the same node, $\tau \in (0,1]$ is a relative threshold, and $N_{\text{min}}$ is an absolute cutoff. 
In short, edges are kept only if they occur frequently relative to the most common activity and exceed a minimum support level, effectively removing spurious detections.

\subsection{Implementation}
We set $\tau = 0.4$, and $N_{\text{min}}=2$. As a general-purpose detector, we used YOLO-World-V2~\cite{yoloworld}; for segmentation, SAM2~\cite{sam}; for head localization, SSD MobileNet~\cite{howard2017mobilenet}; for 3D head orientation estimation, 6DRepNet360~\cite{sixdrepnet360}; and for both VLM and LLM components, GPT-5~\cite{singh2025openai}. The same model configuration was employed throughout all runs.

\section{Experiments}
Our experiments quantitatively evaluate the accuracy of \modelname{} in predicting human activities in 3D environments, its applicability to social scene understanding, and qualitatively assess its use in a real-world application.
We rely on our newly created benchmark, which includes human activities and is designed to assess models on human-centric reasoning and queries. 
We finally showcase its application in a navigation task.
\subsection{\datasetname{}~Benchmark}
As discussed in Sec.~\ref{sec:related_work_3dsg}, existing 3DSG datasets focus on environments without humans, leaving a gap in representing people and their interactions. 
To address this, we introduce the \emph{\datasetname{}} benchmark, comprising 8 synthetic home scenes built in Unity and 3 in the real world, collected with the Odin1 sensor, each with 3 to 9 people interacting with one another and with objects. 
The humans here are static but display facial movements that allow the detection of activities such as speaking.
For each scene, we provide ground-truth point clouds and registered RGB-D images, each paired with a ground-truth semantic mask of the objects in the scene.
In addition, every scene includes a file listing all relationships among human instances and containing queries about the people in the scene, along with the corresponding instance IDs that satisfy them. 
The set of relationships observed in the real scene fully encompasses those present in the synthetic scene.

The relations were annotated by nine trained annotators following a predefined protocol.
For each sequence, annotators were provided with the robot's RGB video from its egocentric camera and an interactive 3D visualization of the reconstructed scene with object-level models and poses.
The multi-view setup was designed to reduce ambiguity caused by occlusions and depth uncertainty.
For each human in the scene, annotators independently described the applicable relation in an open-vocabulary manner based on the image sequence. 
Final labels were first standardized using VerbAtlas \cite{di2019verbatlas} and were retained if at least five of the nine annotators agreed on the same action for a given human--entity pair.
Benchmark details are in Tab.~\ref{tab:dataset_aggregate}; scene queries are detailed in Sec.~\ref{sec:semantic_scene_understanding}

\begin{table}[t!]
\centering
\small %
\begin{tabular*}{\columnwidth}{@{\extracolsep{\fill}}lr}
    \toprule
    \textbf{Information}                         & \textbf{Count} \\
    \midrule
    Total Unique Humans                     & 54               \\
    Total Relationships (GT)                & 146            \\
    \addlinespace %
    
    \multicolumn{2}{@{}l}{\textbf{Query Distribution}} \\
    \hspace{1em}Total Activity Queries                  & 38              \\
    \hspace{1em}Total Functional Queries                & 26              \\
    \hspace{1em}Total Spatial Queries                   & 16              \\
    \addlinespace
    
    \multicolumn{2}{@{}l}{\textbf{Activity Type Distribution}} \\
    \hspace{1em}Total Unique Activity Types             & 11              \\
    \midrule
    \multicolumn{2}{@{}p{\dimexpr\columnwidth-2\tabcolsep}}{\footnotesize \textbf{Activities:} \textit{COOK, INTERACT, LIE, LISTEN, READ, REST, SEE, SIT, SPEAK, STAND, USE}} \\
    \bottomrule
\end{tabular*}
\caption{Aggregate information for the \datasetname{} benchmark. All the relationship frames used are reported.}
\label{tab:dataset_aggregate}
\vspace{-16pt} 
\end{table}

\subsection{Activity Relationships Prediction}\label{sec:activity_relationships_prediction}
In the following, we describe how we evaluated \modelname{} against two baselines for the human relationships prediction on our benchmark.
As the first baseline, we utilized \cgfullname{} (\cgname{}), our extension of ConceptGraphs for detecting human activity, enabling the VLM prompt to relate humans with the other entities. 
Specifically, \cgname{} follows the original ConceptGraphs pipeline except for its VLM prompting stage: humans are added as nodes alongside objects, and the VLM is prompted with the same marker-annotated image used by \modelname{}'s Activity Descriptor to propose relationships between node pairs co-occurring in that single frame; entities outside the frame are never linked, and frames are processed independently with no cross-frame aggregation. Unlike \modelname{}, \cgname{} has no mechanism to estimate head pose or visual field, reason about occlusions, or validate candidates outside the current frame, and thus cannot confirm the remote activities that the Interaction Context Estimator and Activity Solver are designed to capture.
As a second baseline, we consider \modelname{} without generating the behavior description in a separate step, but rather together with the step of proposing relationships, which we refer to as \ablation{}.
As input to \modelname{} and \ablation{}, we use our benchmark registered RGB-D images and a 3DSG generated by the standard (without humans) ConceptGraphs pipeline. 
\cgname{} input is the registered RGB-D images.
We evaluated the performance of all three models in predicting human activities by comparing their outputs with the GT using precision, recall, and F1-score metrics. 
Following the literature \cite{conceptgraph, koch2024open3dsgopenvocabulary3dscene}, a prediction was considered a true positive if it had at least 10\% intersection-over-union (IoU) with the ground-truth point clouds of the related objects, matched the ground-truth label, and its relationship label belonged to the same semantic frame~\cite{di2019verbatlas}. 
Results are reported in Tab.~\ref{tab:aggregate_comparison}. 
We observe a performance gap between synthetic scenes (Maps 1–8) and real-world environments (Maps 9–11); specifically, metrics for real scenes are consistently lower across all models, as sensor and re-projection noise contribute to the prediction error.
\begin{table}[t!]
    \centering
    \setlength{\tabcolsep}{4.2pt}
    \rowcolors{3}{gray!10}{white}
    \begin{tabular}{c|ccc|ccc|ccc} %
        \toprule
        \multirow{2}{*}[-0.6ex]{\textbf{Map}} 
        & \multicolumn{3}{c|}{\textbf{\cgname{}}} 
        & \multicolumn{3}{c|}{\textbf{\modelname{}$_{\text{w/o BD}}$}} 
        & \multicolumn{3}{c}{\textbf{\modelname{} (ours)}} \\ %
        \cmidrule(lr){2-4} \cmidrule(lr){5-7} \cmidrule(lr){8-10}
        \rowcolor{white}
        & P & R & F1 & P & R & F1 & P & R & F1 \\
        \midrule
        Map1  & \textbf{75.0} & 27.3 & 40.0 & 50.0 & 29.3 & 36.9 & 56.3 & \textbf{81.8} & \textbf{66.7} \\
        Map2  & \textbf{65.0} & 10.7 & 18.4 & 30.0 & 14.7 & 19.7 & 58.1 & \textbf{64.3} & \textbf{61.0} \\
        Map3  & 60.0 & 33.3 & 42.8 & \textbf{75.0} & 37.5 & 50.0 & 58.3 & \textbf{77.8} & \textbf{66.7} \\
        Map4  & \textbf{50.0} & 37.5 & 42.9 & 46.7 & 75.0 & 57.6 & 46.7 & \textbf{87.5} & \textbf{60.9} \\
        Map5  & 33.3 & 25.0 & 28.6 & \textbf{70.0} & 25.0 & 36.8 & 50.0 & \textbf{75.0} & \textbf{60.0} \\
        Map6  & 40.0 & 11.1 & 17.4 & 45.5 & 56.0 & 50.2 & \textbf{65.0} & \textbf{72.2} & \textbf{68.4} \\
        Map7  & \textbf{66.7} & 40.0 & 50.0 & 33.5 & 43.5 & 37.9 & 47.5 & \textbf{60.0} & \textbf{53.0} \\
        Map8  & \textbf{77.5} & 41.2 & 53.8 & 28.6 & 41.8 & 34.0 & 64.3 & \textbf{52.9} & \textbf{58.0} \\
        Map9  & \textbf{47.3} & 20.4 & 28.5 & 20.6 & 30.3 & 24.5 & 39.2 & \textbf{63.5} & \textbf{48.5} \\
        Map10 & \textbf{52.0} & 23.2 & 32.1 & 33.7 & 40.2 & 36.7 & 40.5 & \textbf{60.3} & \textbf{48.5} \\
        Map11 & \textbf{56.0} & 10.0 & 17.0 & 36.5 & 36.7 & 36.6 & 43.2 & \textbf{55.6} & \textbf{48.6} \\
        \midrule
        Total & \textbf{56.6} & 25.4 & 33.8 & 42.7 & 39.1 & 38.3 & 51.7 & \textbf{68.3} & \textbf{58.2} \\
        \bottomrule
    \end{tabular}
    \caption{Relationship prediction results on all the \datasetname{} maps. Reported as percentages of Precision (P), Recall (R), and F1-Score (F1).}
    \label{tab:aggregate_comparison}
    \vspace{-8pt}
\end{table}

\begin{table}[!t]
\centering
\footnotesize
\renewcommand{\arraystretch}{1.1}
\setlength{\tabcolsep}{6pt}
\rowcolors{3}{gray!10}{white}
    \begin{tabular}{l|c|c}
\toprule
\textbf{Method} 
        & \textbf{\begin{tabular}{c}Inference (s/frame) \end{tabular}}
        & \textbf{\begin{tabular}{c}Performance (F1)\end{tabular}} \\
        \midrule
        CGH                & 3.24 $\pm$ 0.67 & 33.8  \\
        \modelname{}      & 5.43 $\pm$ 0.78 & \textbf{58.2}  \\
        \modelname{}$_{\text{w/o BD}}$ & 4.96 $\pm$ 0.73 & 39.9  \\
        Local\modelname{}        &  \textbf{2.60 $\pm$ 0.56} & 45.1  \\
        \bottomrule
\end{tabular}
\caption{Comparison of relationship prediction performance and inference time reported as mean $\pm$ standard deviation per frame with at least one detected human.
}
\label{tab:inferencetime}
\vspace{-16pt}
\end{table}

Although \modelname{} achieves the highest overall F1-score, it does not outperform across all metrics. Indeed, \cgname{} achieves higher precision, but this advantage stems from a conservative prediction style that focuses on only a few high-confidence activities while overlooking many valid interactions. 
Such behavior highlights the model's lack of a mechanism to identify a broader range of relationships beyond the single-image frame.
\ablation{} predicts a larger activity space, but its performance is consistently weaker than \modelname{}. 
In particular, \ablation{} struggles with activities requiring contextual reasoning beyond immediate visual cues.
For example, in a cluttered scene where a person’s direct line of sight to an object is partially obscured, \modelname{} can still infer a \texttt{USE} relationship by leveraging the broader context encoded in the behavior description. In contrast, \ablation{} is more likely to miss such subtle interactions, as it loses important details about the person’s focus and intent when processing a cluttered annotated image in a single pass.
This illustrates that the single-step behavior description is a core component of our method, as it helps the system mitigate hallucinations and identify distant activities. 

However, the superior reasoning of \modelname{} comes at the cost of double inference, creating a trade-off between accuracy and speed. 
We report F1 performance relative to inference time in Tab.~\ref{tab:inferencetime}. 
To address this, we also benchmark \modelname{} using Qwen3-VL-8B-Instruct \cite{qwen2025}: a lighter, locally hosted VLM as an alternative to GPT-5. 
This configuration is reported under the name Local\modelname{}. Our results suggest that by utilizing this more efficient architecture and avoiding expensive remote API calls, Local\modelname{} can improve performance over the previous state-of-the-art while also maintaining significantly higher speeds.

\subsection{Social Scene Understanding}\label{sec:semantic_scene_understanding}

\begin{figure}[t]
    \centering
    \includegraphics[width=0.9\linewidth]{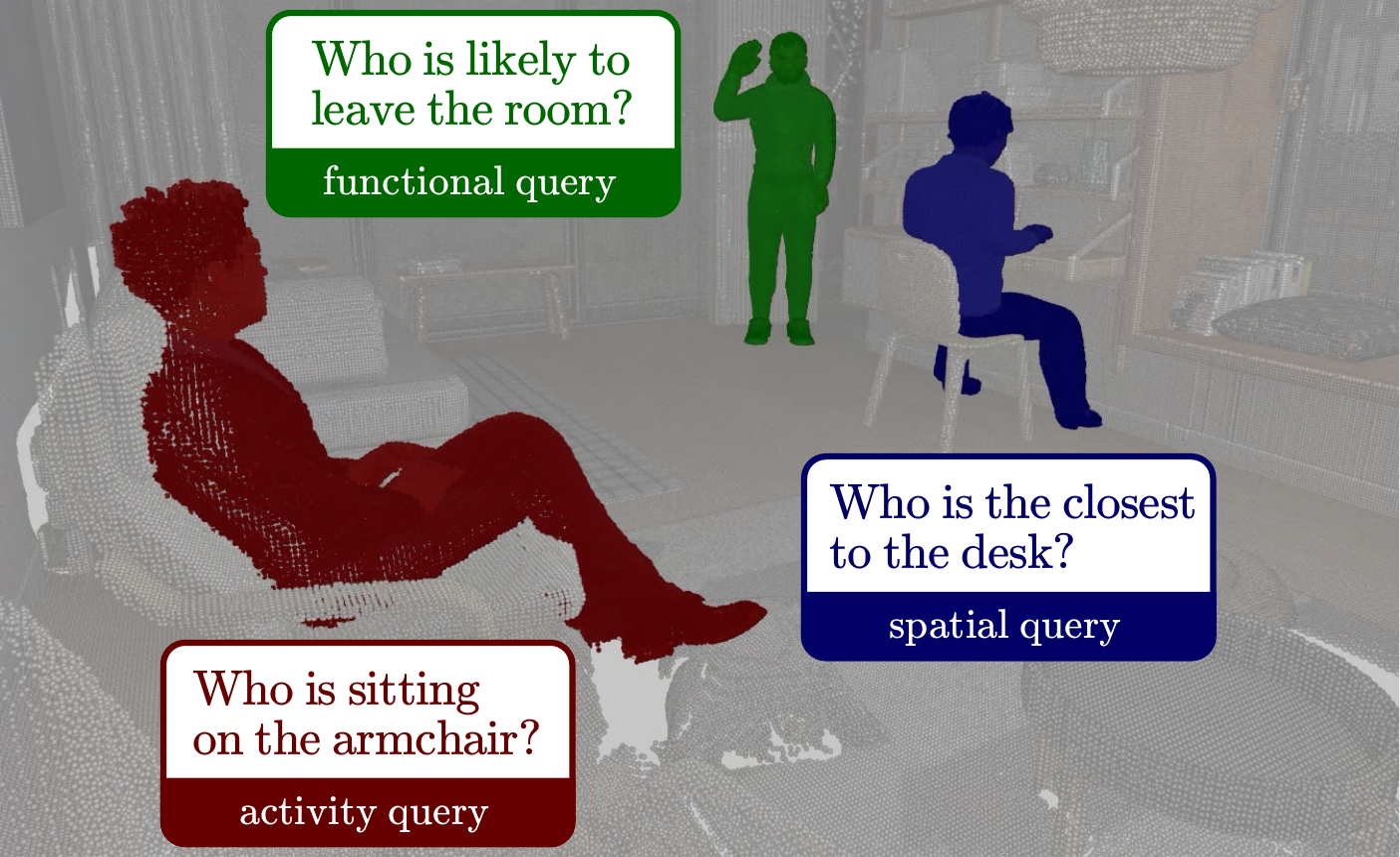}
    \caption{Illustrative example from our \reasoningtask benchmark: queries and their corresponding 3D answers from our method are shown in the same color. From top to bottom, their type is \textit{functional} (green), \textit{spatial} (blue), and \textit{activity} (red).}
    \label{fig:affordance}
    \vspace{-16pt}
\end{figure}

We designed a set of questions to evaluate the ability of our representation to support 3D social scene understanding, i.e., given an input 3D scene representation and a natural language query about the scene involving human activities and interactions, the task is to detect the corresponding human instance in the representation (see Fig.~\ref{fig:affordance}).
The queries are divided into three categories of increasing complexity:

\begin{itemize}
\item \textbf{Spatial queries} focus on spatial relationships between humans and objects, such as ``Who is the closest to the lamp?” or ``Who is next to the window?”.
\item \textbf{Activity queries} involve combining spatial relationships with human activity understanding, such as ``Who is watching TV on the bed?”.
\item \textbf{Functional queries} involve spatial relationships, human activities, and commonsense reasoning, such as ``Who is likely to want to eat some food?”. Answering such questions requires the model to infer, for example, that a person sitting at the table while watching someone cook is likely to want to eat soon.
\end{itemize}
We have a total of 80 queries across 8 scenes, with 105 points obtainable in total, one for each correct instance retrieved.
For each query, the correct solution contains at most two instances, as we are also interested in capturing activities between pairs of people.
As query complexity increases, the solution must integrate more information from the Social 3D Scene Graph and perform more advanced reasoning.
We evaluate the ability to retrieve the correct person, given a query, using two different feature-based baselines: CLIP\_ALL, which compares the CLIP \cite{clip} features of the query with those of all the objects in the graph (as is standard in 3DSG sota methods \cite{hov-sg, conceptgraph, clio}), and outputs the most likely match. 
To reduce the search space and assess how much can be captured with purely visual observations, we also use CLIP\_FAV, which follows the same CLIP approach but restricts the comparison to the multimodal features of human nodes only. 

\begin{table}[!t]
\centering
\footnotesize
\renewcommand{\arraystretch}{1.1}
\setlength{\tabcolsep}{6pt}
\rowcolors{3}{gray!10}{white}
\begin{tabular}{l|c|c|c|c|c}
\toprule
\textbf{Method} & \textbf{Spatial} & \textbf{Activity} & \textbf{Functional} & \textbf{Mean} & \textbf{Pts} \\
\midrule
\cgname{}        & 46.4 & 44.6 & 41.5 & 44.2 & 47 \\
CLIP\_FAV  & \underline{64.3} & \underline{52.1} & \underline{55.1} & \underline{57.2} & 57 \\
CLIP\_ALL & 31.0 & 24.6 & 25.1 & 26.9 & 29 \\
\textbf{\modelname{} (ours)}     & \textbf{79.8} & \textbf{81.9} & \textbf{64.6} & \textbf{75.4} & \textbf{81} \\
\bottomrule
\end{tabular}
\caption{Social scene understanding results per query type. Reported as the ratio (\%) between the collected points (Pts) and the total points. 
}
\label{tab:suntest}
\vspace{-16pt}
\end{table}

We moreover compare our method with the LLM reasoning approach of ConceptGraphs, adapted to our CGH output. This solution consists of feeding the LLM with a detailed JSON description of the nodes and the relationships between them. 
We find this approach inefficient and contrary to the principle of maintaining a lightweight representation that the LLM can parse effectively in order to perform the complex reasoning required to answer functional queries. 
For this reason, we developed our own JSON format of the Social 3DSG, which is then provided to the LLM together with the query.
Each node is represented as \texttt{[id, class, 3D center]}, and each edge as \texttt{[id\_from, id\_to, relationship\_frame]}, thereby removing the additional edge descriptions and object tags used in the baselines.
However, our representation does not lose this level of detail, since we also associate with the graph a compact JSON containing the frame description, gloss, and example actions provided by the inventory \cite{di2019verbatlas} (e.g., frame name, gloss, and example verbs) for each unique relationship frame in the graph.
For feature-based methods, we retrieve the top-2 most likely nodes if the query requires two results, while for the LLM parsing approach, we compare the GT with at most the top-2 candidate results. 
We assign one point for each correctly retrieved human in the GT if the retrieved method has an IoU $\geq 0.10$, and then compute the ratio between obtained points and total possible points. 
We report the results by query type in Tab.~\ref{tab:suntest}.

This experiment makes it clear that, unlike in object-based queries, feature-based solutions are not well-suited for processing large 3DSGs to handle complex human-centered queries. 
They perform the worst, mainly because the bag-of-words effect often shifts the focus toward the objects mentioned in the query, even when only a small number of humans are involved. 
They can find the correct answer when the object mentioned in the query is close to the mask into which it has been embedded.
This occurs most often for spatial queries, but when the task requires processing more remote relationships, their limitations become evident.
From the scores, we also observe that our more compact representation not only helps in processing activity relationships required to answer queries, but also enables more effective reasoning for spatial queries. By reducing the number of tokens, it prevents context window overflow that can cause LLMs to lose track of relevant information, thereby demonstrating the superiority of our serialization.

\subsection{Application in a Downstream Task}
As a final experiment to qualitatively validate the practical utility of \modelname{} in a real-world downstream task, we conducted a navigation experiment in a novel scene with RGB-D data captured using an iPhone 15 Pro (30 FPS, downsampled to 5 FPS). The scene was staged to depict a common social situation: two people engaged in conversation.

\begin{figure}[t]
    \centering
    \includegraphics[width=0.9\linewidth]{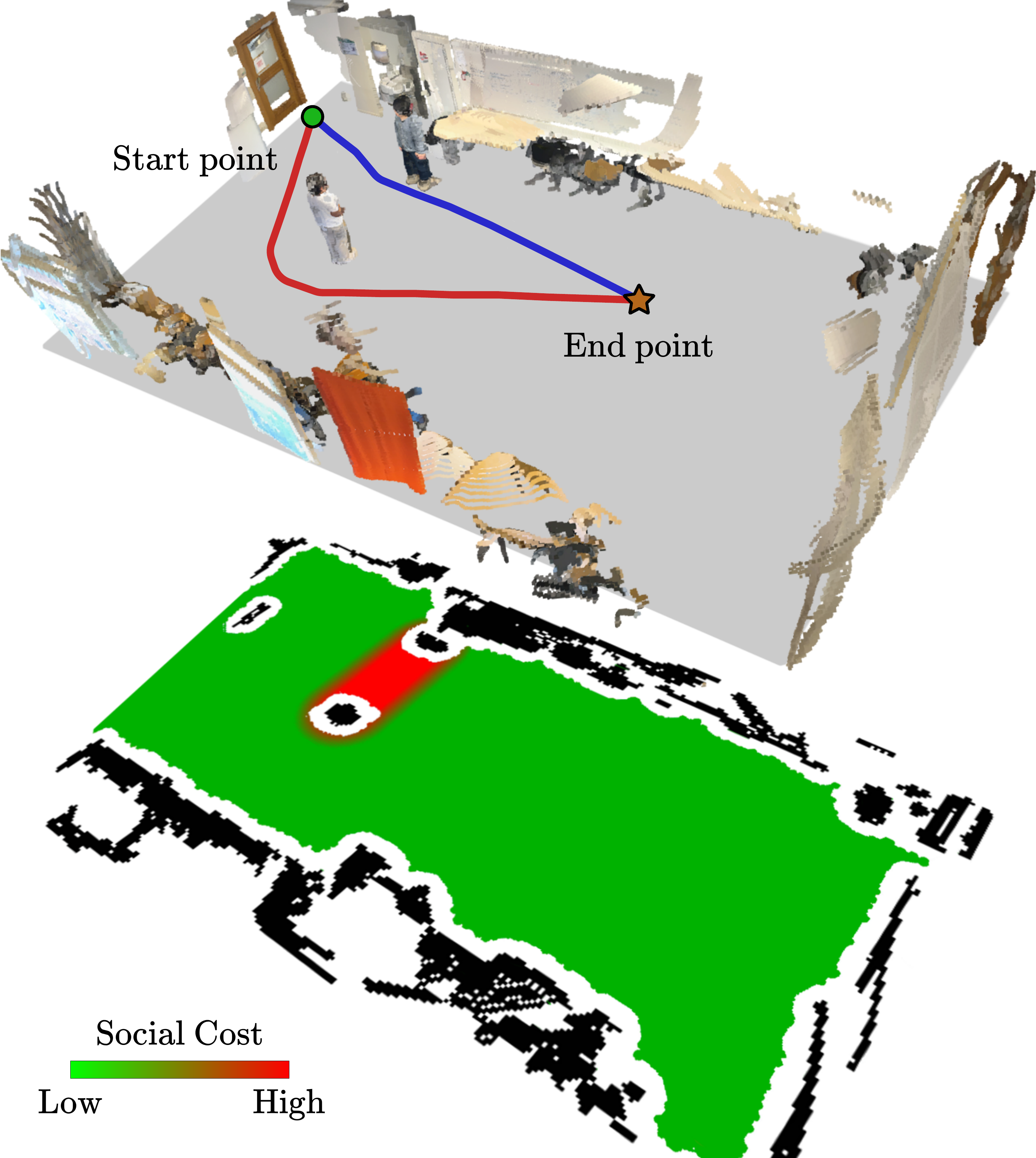}
    \caption{Socially aware planning in a real-world scene based on the Social Scene 3D Graph obtained by \modelname{}. Without social awareness, the planner generates a trajectory that passes between two people engaged in conversation (blue), whereas incorporating social cost produces a trajectory that is longer but avoids interrupting the interaction (red).}
    \label{fig:real_nav}
    \vspace{-16pt}
\end{figure}

We integrate \modelname{} into a 3DSG pipeline, generating both a 3D instance-segmented point cloud of the environment and the corresponding Social 3D Scene Graph. %
This reconstructed map was then provided to a sample-based motion planner~\cite {9158399}, which produced a conventionally cost-optimal (shortest) trajectory from a start to a goal position in real time, visualized as the blue path in Fig.~\ref{fig:real_nav}.
Next, we leverage the Social 3D Scene Graph for socially-aware motion planning by augmenting the geometric cost map with a ``social cost''.
Specifically, interactions such as the identified \texttt{SPEAK} relationship create a high social-cost region between the two individuals (visualized in the lower layer of Fig.~\ref{fig:real_nav}). 
This field penalizes paths that intersect the interaction space, with the cost at any point $p$ calculated as:
\begin{equation}
C_{\text{social}}(p) =
\begin{cases}
C_{\text{rel}} \times \left( 1 - \frac{d(p, L_{ij})}{R} \right) & \text{if } d(p, L_{ij}) < R \\
0 & \text{otherwise}
\end{cases}
\end{equation}
where $C_{\text{rel}}$ is a cost autonomously derived from the relationship type, $R$ is its radius of influence, and $d(p, L_{ij})$ is the distance between point $p$ and the line segment $L_{ij}$ connecting the two individuals.
Presented with this socially augmented cost map, the planner generates a new trajectory (the red path) that avoids disrupting social interaction, even if it is slightly longer. 
This shows how \modelname{}'s semantic understanding can generate socially normative behavior, extending prior social-navigation work constrained to simpler scene models~\cite{7487452}.

\section{Limitation}
Although our work establishes a solid foundation for integrating humans and their activities into 3DSGs, several limitations remain.
First, our approach assumes nearly static scenes, and the current inference speed is not yet optimized for the real-time processing required in highly dynamic environments, which were not considered in our study.
Second, the interaction context is currently confined to the person’s immediate field of view. For instance, the model cannot resolve complex behaviors in which a person points to a distant object while looking elsewhere.

\section{Conclusions}
In this work, we introduced Social 3D Scene Graphs, a representation that we envision as a foundation for socially intelligent robots and a method, \modelname,  to build them.
By modeling humans, objects, and their interactions, this structure supports applications ranging from human-aware navigation to 3D social scene understanding.
Through our benchmark, \datasetname, we provide a grounding for evaluating such capabilities and through our experiments, we demonstrate how this representation improves a robot's understanding of the environment.
While our current approach is limited to static scenes, we plan to extend it to dynamic environments by developing ad hoc relationship-extraction models.
Moreover, incorporating other modalities such as audio could provide richer context, e.g., to disambiguate human states like leaving versus arriving. Looking ahead, we believe that by continuously tracking and learning from human behavior in shared spaces, Social 3D Scene Graphs can support long-term personalization, allowing robots to capture individual preferences and proactively anticipate human needs.

\bibliographystyle{IEEEtran}
\bibliography{main}

\end{document}